\documentclass[sigconf]{acmart}

\usepackage{booktabs} 

\usepackage{latexsym}
\usepackage[colorinlistoftodos]{todonotes}
\usepackage{amsmath}
\usepackage{amssymb}
\usepackage{url}
\usepackage{color}

\newcommand{\delete}[1]{} 			

\newcommand{\Ni}{({\em i})~}
\newcommand{\Nii}{({\em ii})~}
\newcommand{\Niii}{({\em iii})~}

\newcommand{\query}{new\,}
\newcommand{\forum}{related\,}
\setcopyright{rightsretained}
\acmDOI{10.475/123_4}

\acmISBN{123-4567-24-567/08/06}

\acmConference{SIGIR '17}{}{August 07-11, 2017, Shinjuku, Tokyo, Japan}
\acmYear{2017}
\copyrightyear{2017}

\acmPrice{15.00}

\begin{document}
\title{Cross-Language Question Re-Ranking}

\author{Giovanni Da San Martino, Salvatore Romeo, Alberto Barr\'on-Cede\~no, Shafiq Joty, Llu\'is M\`arquez, Alessandro Moschitti, Preslav Nakov}
\affiliation{%
\institution{Qatar Computing Research Institute, HBKU}
  \streetaddress{HBKU Research Complex. P.O. Box 5825}
  \city{Doha} 
  \country{Qatar}
}
\email{{gmartino, sromeo, albarron, sjoty, lmarquez, amoschitti, pnakov}@hbku.edu.qa}


\renewcommand{\shortauthors}{Da San Martino et al.}

\begin{abstract}
We study how to find relevant questions in community forums when the language of the \query questions is different from that of the existing questions in the forum. In particular, we explore the Arabic--English language pair. We compare 
a kernel-based system with 
a feed-forward neural network in a scenario where a large parallel corpus is available for training a machine translation system, bilingual dictionaries, and cross-language word embeddings.
We observe that both approaches degrade the performance of the system when working on the translated text, especially the kernel-based system, which depends heavily on a syntactic kernel.  
We address this issue using a cross-language tree kernel, which compares the original Arabic tree to the English trees of the \forum questions. We show that this kernel almost closes the performance gap with respect to the monolingual system. On the neural network side, we use the parallel corpus to train cross-language embeddings, which we then use to represent the Arabic input and the English \forum questions in the same space. 
The results also improve to close to those of the monolingual neural network. Overall, the kernel system shows a better performance compared to the neural network in all cases.
\end{abstract}

\copyrightyear{2017}
\acmYear{2017}
\setcopyright{acmlicensed}
\acmConference{SIGIR '17}{August 07-11, 2017}{Shinjuku, Tokyo,
Japan}

%
%



\maketitle

\newpage
\section{Introduction}




In this paper, we study the problem of question re-ranking, which is an important task of the more general problem of community Question Answering (cQA). In particular, we address question re-ranking in a cross-language (CL) setting, i.e.,~where the language of the \query question is different from the language of the candidate questions.
We explore alternative ways to adapt kernel-based systems for English into this setting, when the query language is Arabic. This is an interesting scenario because state-of-the-art cQA models rely upon relational syntactic/semantic structures, using Tree Kernels (TKs)~\cite{filice-EtAl:2016:SemEval}, 
and these might be difficult to port across translation--based models. We compare the kernel machines to feed-forward neural networks (FNN), which have been known to perform well for cQA~\cite{nakov-marquez-guzman:2016:EMNLP2016}.

We first explore a standard approach in CLIR: translating the input questions and applying our monolingual systems on the English-translated text. 
Our second approach, which is novel, is based on a CL TK ---which does not require any translation--- as it is applied directly to pairs of Arabic and English trees. This tree kernel makes use of a statistical bilingual dictionary extracted from a parallel corpus. The FNN system can also make use of the parallel corpus by learning cross-language embeddings, which we further use in order to compare the Arabic and the English input representations directly.  

We tested our approaches on the benchmark datasets from the SemEval-2016 task 3 on cQA~\cite{nakov-EtAl:2016:SemEval2}, which we enriched with Arabic \query questions. 
The results show that machine translation does not drastically degrade the ranking performance, probably because of the robustness of our similarity features. Most importantly, the use of the cross-language tree kernels almost fills the gap with respect to the monolingual system.
%

\section{Related Work}

Question re-ranking can be approached from several different angles. \citet{Cao:2008:RQU:1367497.1367509} tackled it by comparing representations based on topic term graphs, i.e., by judging topic similarity and question focus. 
\citet{Jeon:2005:FSQ:1099554.1099572} and \citet{zhou2011phrase} dodged the lexical gap between questions by assessing their similarity on the basis of a (monolingual) translation model. \citet{wang2009syntactic} computed a similarity function on the syntactic-tree representations of the questions. A different approach using topic modeling for question retrieval was introduced by~\citet{ji2012question} and~\citet{zhang2014question}, who used LDA topic modeling to learn the latent semantic topics in order to retrieve similar questions. \citet{dossantos-EtAl:2015:ACL-IJCNLP} used neural networks for the same purpose. 

\noindent Cross-language approaches have mainly focused on Question Answering (QA). This has been fostered by multiple challenges such as 
the Multilingual QA Challenge at CLEF 2008~\cite{clef/FornerPAAFMOPRSSS08}, NTCIR-8's Advanced Cross-lingual Information Access (ACLIA)~\cite{mitamura:10}, and DARPA's Broad Operational Language Technologies (BOLT) IR task~\cite{Soboroff:2016}. 
Usually, the full question is translated using an out-of-the-box system in order to address CL-QA~\cite{Hartrumpf:08,Lin:10}. 
\citet{ture-boschee:2016:EMNLP2016} proposed supervised models to combine different translation settings. 
Some approaches translate only keywords~\cite{Ren:10}. 
To the best of our knowledge, no research has been carried out on CL question 
re-ranking before. 
Regarding cross-language tree kernels, the only previous study relates to mapping natural language to an artificial language (SQL) \cite{Giordani:2009:SMN:2174299.2174323, DBLP:conf/coling/GiordaniM12}. 
We use a similar cross-language tree kernel along with the new idea of deriving relational links \citep{sigir12} using cross-language dictionaries.

\section{Task and Corpora}\label{sec:taskAndCorpora}

We experiment with data from the SemEval-2016 Task 3 on Community Question Answering~\cite{nakov-EtAl:2016:SemEval2}, which we further augment with translations as described below. We focus on subtask B, which targets \emph{question--question similarity} (QS).
Given a \query question $q_o$ 
and the set of ten \forum questions from the QatarLiving forum $q_1, q_2, \dots, q_{10}$, retrieved by a search engine, the goal is to re-rank the \forum questions according to their similarity with respect to the \query question. 
The relationship between $q_o$ and $q_i$, $i \in \{1,2,\ldots,10$\}, is described with a label: \textsc{PerfectMatch}, \textsc{Relevant}, and \textsc{Irrelevant}. The goal is to rank the questions with the first two labels higher than those with the latter label. 
Note that the \emph{questions} in this dataset are generally long multi-sentence stories which are written in informal English; full of typos and ungrammaticalities.
The SemEval data has 267 \query questions for training, 50 for development, and 70 for testing, and ten times as much $q_o$--$q_i$ pairs: 2,670, 500, and 700, respectively.

Based on this data, we simulated a \emph{cross-language cQA setup}. We first got the 387 \query train+dev+test questions translated into Arabic by professional translators.%
\footnote{The extension of the dataset is available at \url{http://alt.qcri.org/resources/cqa}.}
Then, we used these Arabic versions of the questions as input with the goal of re-ranking the ten \forum English questions.

We also used an Arabic--English parallel corpus, which includes the publicly available TED and OPUS corpora~\cite{Tiedemann:12}. We used this corpus in order to train an in-house phrase-based Arabic-English machine translation (MT) system,\footnote{The MT system also uses a language model trained with the English Gigaword.} and also to extract a bilingual dictionary in order to learn cross-language embeddings, as described in Sections~\ref{sec:svm-model} and~\ref{sec:fnn-model} below.

\section{A Kernel-based System} 
\label{sec:svm-model}

We address the re-ranking task described in Section~\ref{sec:taskAndCorpora} by using the scoring function of a binary (\{\textsc{PerfectMatch} $\cup$ \textsc{Relevant}\} vs. \textsc{Irrelevant}) classifier based on support vector machines (SVM): $r(q_o,q_i)=\sum_{j}^ny_j\alpha_jK((q_o,q_i),(q_o^j,q_i^j))$ in order to rank all the \forum questions $q_i$ with respect to their corresponding \query question $q_o$. 
Here $K(\cdot)$ is a kernel function assessing how similar two pairs of questions are. 
We use a combination of kernels on tree pairs and features as described below. 
	
\subsection{Tree Kernels}
Given a pair of syntactic trees of the questions and a kernel for trees $K^T$, we  define the following kernel applied to $(q_o,q_i), (q_{o}^{j}, q_i^j)$:
\begin{equation}
\hspace{-2em}  K^T(t(q_{o},q_i),t(q^j_{o},q^{j}_i))+K^T(t(q_i,q_{o}),t(q^j_{i},q^j_{o})), \label{eq:tkonpairs}\hspace{-.5em}
\end{equation}
where $t(x,x')$ is a string transformation method that returns the parse tree for the text $x$, further enriching it with \emph{REL}ational tags computed with respect to the syntactic tree of $x'$. Typically, REL tags are assigned to the matching words between $x$ and $x'$, and they are propagated to the parent and grand-parent nodes (i.e., up to 2 levels). This kernel in the monolingual setting is described in~\cite{barroncedeno-EtAl:2016:SemEval}. 

Note that Eq.~(\ref{eq:tkonpairs}) can be applied to pairs $(q_{o},q_i)$ in which $q_{o}$ and $q_i$ are texts in different languages, since in Eq.~\eqref{eq:tkonpairs} the \query (resp.~\forum) questions are only compared to \query (resp. \forum) questions: this produces the kernel space of tree fragment pairs as shown in \cite{Giordani:2009:SMN:2174299.2174323, DBLP:conf/coling/GiordaniM12}, where the pair members are in different languages. 
Moreover, the definition of $t(x,x')$ is more complicated in case $x$ and $x'$ are in different languages as, in addition to using separate Arabic and English parsers, we need to define methods for matching words in different languages. Given the rich morphology of  Arabic, this is not a trivial task.
\smallskip

\noindent\textbf{Cross-Language Tree Matching}. 
In order to match the lexical items from both trees, we created a word-level Arabic-to-English statistical dictionary using IBM's Model 1~\cite{Brown:90} 
over our bilingual corpus, which we pre-processed using Farasa \cite{DARWISH16.164} in order to share the segmentation and diacritization of the Arabic syntactic parser.

\subsection{Feature Vectors} 
\label{sec:features}
We combined the above tree kernels linearly with RBF kernels applied to the following four feature vectors:\smallskip

\noindent\emph{ConvKN features.}
We 
used the 21 features proposed in~\cite{barroncedeno-EtAl:2016:SemEval} computing similarities between the \query and \forum questions such as: longest common subsequences, Jaccard coefficient, word containment, cosine similarity. Since such similarities can be only computed when the two texts are in the same language, we use the English translation to obtain them for the cross-language system. 
\smallskip 

\noindent\emph{Embedding features.}
We used three types of vector-based embeddings in order to encode the text of a question:
(1)~\textsc{Google\_vectors}: 300-dimensional embedding vectors, pre-trained on Google News \cite{mikolov-yih-zweig:2013:NAACL-HLT}; 
(2)~\textsc{QL\_vectors}: we trained domain-specific vectors using \textsc{word2vec} on all available QatarLiving data, both annotated and raw (as provided for SemEval-2016 Task 3).
(3)~\textsc{Syntax}: we parsed the questions using the Stanford neural parser~\cite{socher-EtAl:2013:ACL2013}, and we used the final 25-dimensional vector that is produced internally as a by-product of parsing.
We did not use the embeddings themselves, but the cosines between the embeddings of a \query and of a \forum question.
\smallskip

\noindent\emph{MTE features.}
We used the following MT evaluation metrics, which compare the similarity between the \query and a \forum question as in~\cite{ACL2016:MTE-NN-cQA}:
(1)~\textsc{Bleu}; 
(2)~\textsc{NIST}; 
(3)~\textsc{TER} v0.7.25; 
(4)~\textsc{Meteor} v1.4; 
(5)~Unigram ~\textsc{Precision}; 
(6)~Unigram ~\textsc{Recall}.
We further used various components involved in the computation of \textsc{Bleu}, as features:
$n$-gram precisions,
$n$-gram matches,
total number of $n$-grams ($n$=1,2,3,4),
lengths of the \forum and of the \query questions, 
length ratio between them,
and \textsc{Bleu}'s brevity penalty.
\smallskip

\noindent\emph{Task-specific features.}
We computed various 
task-specific features, most of them introduced in the \nobreak{SemEval-2015} Task 3 on cQA~\cite{nicosia-EtAl:2015:SemEval}. This includes some question-level features:
(1)~number of URLs/ images/emails/phone numbers;
(2)~number of tokens/sentences;
(3)~average number of tokens;
(4)~type/token ratio;
(5)~number of nouns/verbs/adjectives/adverbs/ pronouns;
(6)~number of positive/negative smileys;
(7)~number of single/double/triple exclamation/interrogation symbols;
(8)~number of interrogative sentences (based on parsing);
(9)~number of words that are not in \textsc{word2vec}'s Google News vocabulary.
Also, some question-question pair features:
(10)~count ratio in terms of sen\-tences/tokens/nouns/verbs/ adjectives/adverbs/pronouns;
(11)~count ratio of words that are not in \textsc{word2vec}'s Google News vocabulary.
Finally, we also have one meta feature:
(12)~reciprocal rank of the \forum question.

\section{A Neural Network System} 
\label{sec:fnn-model}
\begin{figure}[t]
\includegraphics[width=0.85\columnwidth]{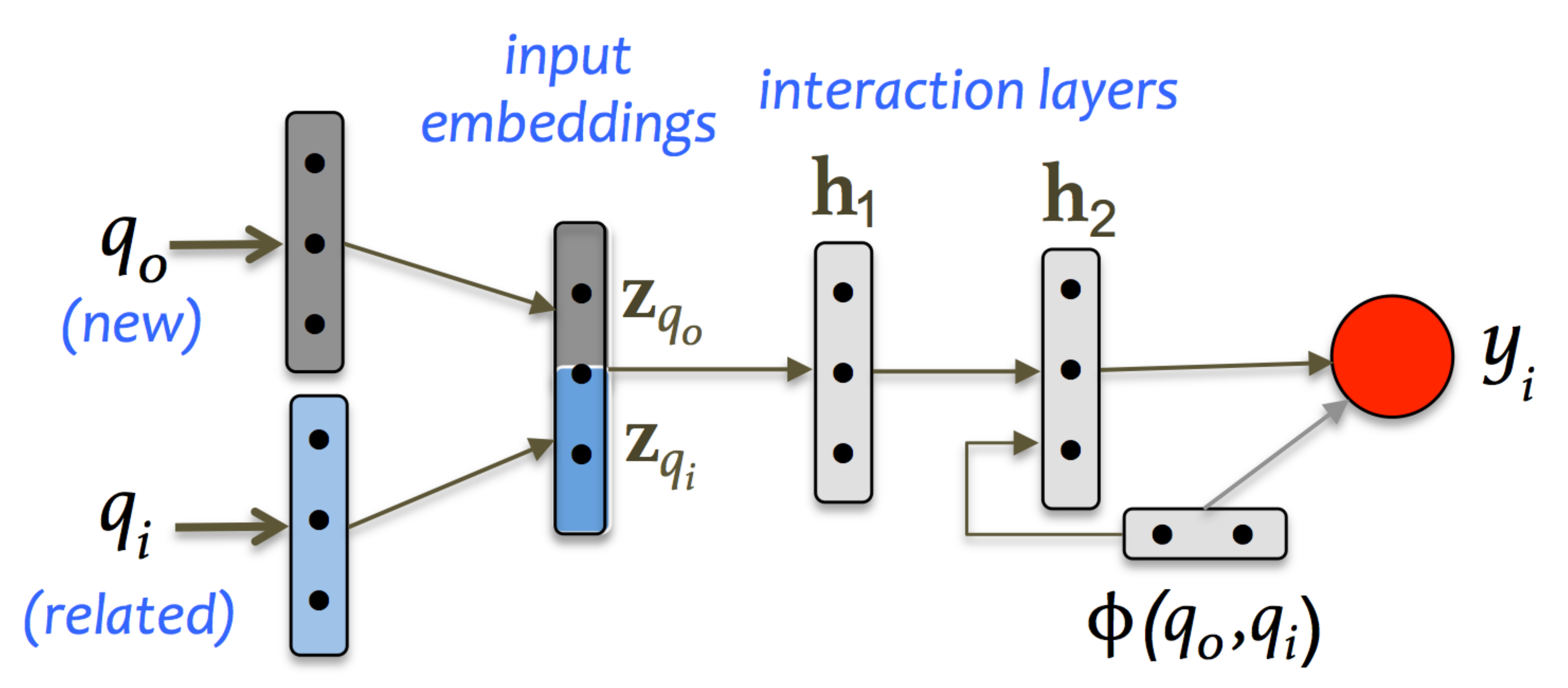}
\vspace{-2mm}
\caption{Feed-forward neural network for QS.}
\label{fig:fnn}
\vspace{-1em}
\end{figure}

Given the small size of the training set, we used a simple Feed-forward Neural Network (FNN), depicted in Figure~\ref{fig:fnn}.
%
The input is a pair $(q_o, q_i)$. We map the input elements to fixed-length vectors $(\mathbf{z}_{q_o}, \mathbf{z}_{q_i})$ using their \emph{syntactic} and \emph{semantic embeddings} (described in Section~\ref{sec:features}). The network then models the interactions between the input embeddings by passing them through two non-linear hidden layers (rectified linear units, ReLU). Additionally, the network also considers \emph{pairwise} features $\phi(q_o,q_i)$ between the two input elements that go directly to the output layer, and also through the second hidden layer. In our case, $\phi(q_o,q_i)$ is the concatenation of the MTE and the task-specific features described in Section~\ref{sec:features}, which are also used by the kernel-based system. 
The following equations describe the transformation: $\mathbf{h}_1 =  f(U [\mathbf{z}_{q_o},\mathbf{z}_{q_i}])$; $\mathbf{h}_2 =  f(V [\mathbf{h}_{1},\phi(q_o, q_i)])$, where $U$ and $V$ are the weight matrices in the first and in the second hidden layer.

The output layer of the neural network computes a sigmoid on the output layer weights and on the pairwise features in order to determine whether $q_i$ is relevant with respect to the \query question $q_o$. We train the models by minimizing the cross-entropy between the predicted distributions and the target distributions, i.e., the gold labels.
\smallskip

\noindent\textbf{Cross-language embeddings}.
\label{subsec:xl-embeddings}
Using our parallel Arabic--English corpus, we trained cross-language embeddings using the bivec method by \citet{luong-pham-manning:2015:VSM-NLP}, a bilingual extension of word2vec, which has reported excellent results on semantic tasks close to ours~\cite{upadhyay-EtAl:2016:P16-1}. Using these CL embeddings allows us to compare directly representations of the Arabic $q_o$ and the English $q_i$ input questions. In particular, we trained 200-dimensional word embeddings using the parameters described in~\cite{upadhyay-EtAl:2016:P16-1}, with a context window of size five and iterating for five epochs.

\section{Experiments}


We consider three scenarios: (\emph{i})~\textbf{Original}, i.e., the SemEval 2016 setup,  (\emph{ii})~\textbf{Translated}, in which $q_o$ are originally in Arabic and machine-translated into English, and  (\emph{iii})~\textbf{Arabic}, in which $q_o$ are in Arabic. In all settings, we apply the tree kernel in Eq.~\eqref{eq:tkonpairs}.
However, we distinguish when the kernel is applied to the original English trees $q_o$ (TK), the translated ones (TK$_{MT}$), and the Arabic ones (TK$_{CL}$).
In all experiments, we kept the default parameters for the kernels and we selected the $C$ parameter of SVM on the development set, trying \{0.01, 0.1, 1, 10\}. 
For the FNN models, we used the development set for early stopping based on MAP as well as for parameter optimization.

\begin{table}[t]
\centering
\begin{tabular}{|r@{ }l|l|cc|}
\hline
	&			& \query		&	\multicolumn{2}{c|}{MAP}\\
	&	system	& question &	dev.			&	test\\\hline
	\hline
1.	&	IR rank				&	English		&	\emph{71.35}	& \emph{74.75} \\
2.	&	UH-PRHLT (SemEval) 			&	English		&	---		& 76.70 \\
3.	&	ConvKN (SemEval)		&	English		&	---		& 76.02 \\ 
4.	&	SVM + TK			&	English		&	73.02	& \textbf{77.41} \\
5.	&	FNN 		&	English		&	72.52	& 76.26 \\\hline\hline
6.	&	SVM + TK$_{MT}$	&	Translated		&	72.94	& 76.67 \\
7.	&	SVM 		&	Translated		&	71.99	& 76.36 \\
8.	&	FNN 		&	Translated		&	72.44	& 75.73 \\\hline\hline
9.	&	SVM + TK$_{CL}$	&	Arabic			&	73.34	& \textbf{77.14}\\
10.	&	FNN + CL\_emb	&	Arabic			&	72.27	& 76.06 \\\hline
   \end{tabular}
   \caption{\label{t-results} MAP scores on the development and test datasets.}
   \vspace{-2em}
\end{table}

Table~\ref{t-results} shows Mean Average Precision (MAP) on the development and on the test datasets.
The first block contains the reference results on the original English test set (rows~1--5). \emph{IR rank} corresponds to the 
Google-generated ranking, which is a hard-to-beat \emph{baseline}. \emph{UH-PRHLT} and \emph{ConvKN} are the two best-performing systems at SemEval-2016 Task 3 (see~\cite{nakov-EtAl:2016:SemEval2} for details). 
\emph{SVM + TK} is the kernel-based system presented in this paper, which reproduces \emph{ConvKN} and adds our extra features (cf. Section~\ref{sec:features}). Finally, \emph{FNN} is the neural network model presented in this paper.

Our SVM model (row 4) shows a sizable improvement over \nobreak{\emph{ConvKN}} on the test set, which means that our extra features are strong. Actually, the SVM results are also better than the best 
system at SemEval-2016 (+0.71 MAP). 
The FNN model shows also competitive performance, but below the SVM system (-1.15 MAP). The monolingual result from SVM (77.41 MAP) is the upper bound performance when considering the results in the CL scenario. 

The second block (rows 6--8) shows the results of our systems in the ``translated'' setting.  One concern about applying the TK approach in this setting was that the translated text might be grammatically broken and the parser could produce low-quality parse trees.
Still, the SVM system degrades its performance only to 76.67 MAP when applied after machine-translating the Arabic \query questions (i.e., -0.74 MAP points below the upper bound). 

Row 7 shows the result for SVM without TK. Its MAP score is 76.36, which is slightly below the previous score of 76.67 obtained with TK; 
this shows that the two kernels provide complementary information. 
Comparatively, the FNN model degrades the performance even less when working with the translated Arabic query (75.73, row 8 vs. 76.26, row 5). This indicates again that the features we use are robust to translation.  

The last block in the table (rows 9-10) shows the results of the systems using the CL kernels and representations. 
The SVM system scores 77.14 MAP when using the CL kernel. This is above the results with TK$_{MT}$. 
This final MAP value is very close to the upper bound system (77.41). In conclusion, achieving a similar ranking quality to that of the monolingual setting is possible by departing from Arabic text and using the novel cross-language tree kernel together with a robust feature set computed on the translated texts.
Finally, the FNN system achieves slightly improved results when we add the input representation based on cross-language embeddings (row 10), reaching a MAP score, 76.06, that is very close to the monolingual FNN (76.26). 

\section{Conclusions and Future Work}
We studied the task of cross-language question re-ranking in community question answering.
We first explored the possibility of using MT for translating an Arabic query question and then applying an English monolingual system. The results of two alternative systems for question re-ranking (kernel- and FNN-based) show a relatively small degradation in performance with respect to the monolingual setting on a well-established SemEval dataset. 
Furthermore, we showed that the performance gap in the SVM system can be almost closed by using a novel cross-language tree kernel, which compares directly the source and the target language trees. 
A cross-language input representation can also help the FNN system to close the gap  with respect to the monolingual case. Finally, the performance of the SVM system is always superior to that of the FNN system in our setting. We conjecture that this is due to the relatively small size of the training set, and due to the information provided by the tree kernel (relations between syntactic sub-structures).

Our work enables interesting future research lines, e.g., \Ni designing more accurate cross-language TKs using better Arabic structures and cross-language word matching and embeddings, \Nii combining the SVM and the FNN models, and \Niii exploring how far a system can go without machine translation.


\begin{acks}
This research was performed by the Arabic Language Technologies group at Qatar Computing Research Institute, HBKU\@, within the Interactive sYstems for Answer Search project ({\sc Iyas}).
\end{acks}

\bibliographystyle{ACM-Reference-Format}
\bibliography{references} 


\begin{thebibliography}{00}


\ifx \showCODEN    \undefined \def \showCODEN     #1{\unskip}     \fi
\ifx \showDOI      \undefined \def \showDOI       #1{{\tt DOI:}\penalty0{#1}\ }
  \fi
\ifx \showISBNx    \undefined \def \showISBNx     #1{\unskip}     \fi
\ifx \showISBNxiii \undefined \def \showISBNxiii  #1{\unskip}     \fi
\ifx \showISSN     \undefined \def \showISSN      #1{\unskip}     \fi
\ifx \showLCCN     \undefined \def \showLCCN      #1{\unskip}     \fi
\ifx \shownote     \undefined \def \shownote      #1{#1}          \fi
\ifx \showarticletitle \undefined \def \showarticletitle #1{#1}   \fi
\ifx \showURL      \undefined \def \showURL       #1{#1}          \fi
\providecommand\bibfield[2]{#2}
\providecommand\bibinfo[2]{#2}
\providecommand\natexlab[1]{#1}
\providecommand\showeprint[2][]{arXiv:#2}

\bibitem[\protect\citeauthoryear{??}{cle}{2008}]%
        {clef:08}
 \bibinfo{year}{2008}\natexlab{}.
\newblock \bibinfo{booktitle}{{\em {Evaluating Systems for Multilingual and
  Multimodal Information Access, 9th Workshop of CLEF 2008, Revised Selected
  Papers}}}. \bibinfo{address}{Aarhus, Denmark}.
\newblock


\bibitem[\protect\citeauthoryear{??}{ntc}{2010}]%
        {ntcir:8}
 \bibinfo{year}{2010}\natexlab{}.
\newblock \bibinfo{booktitle}{{\em {Proc. of NTCIR-8 Workshop Meeting}}}.
  \bibinfo{address}{Tokyo, Japan}.
\newblock


\bibitem[\protect\citeauthoryear{??}{sem}{2016}]%
        {semeval:16}
 \bibinfo{year}{2016}\natexlab{}.
\newblock \bibinfo{booktitle}{{\em Proc. of the 10th SemEval Workshop}}.
  \bibinfo{address}{San Diego, California, USA}.
\newblock


\bibitem[\protect\citeauthoryear{Barr\'{o}n-Cede\~{n}o, Da~San~Martino, Joty,
  Moschitti, {Al-Obaidli}, Romeo, Tymoshenko, and Uva}{Barr\'{o}n-Cede\~{n}o
  et~al\mbox{.}}{2016}]%
        {barroncedeno-EtAl:2016:SemEval}
\bibfield{author}{\bibinfo{person}{Alberto Barr\'{o}n-Cede\~{n}o},
  \bibinfo{person}{Giovanni Da~San~Martino}, \bibinfo{person}{Shafiq Joty},
  \bibinfo{person}{Alessandro Moschitti}, \bibinfo{person}{Fahad {Al-Obaidli}},
  \bibinfo{person}{Salvatore Romeo}, \bibinfo{person}{Kateryna Tymoshenko},
  {and} \bibinfo{person}{Antonio Uva}.} \bibinfo{year}{2016}\natexlab{}.
\newblock \showarticletitle{{ConvKN at SemEval-2016 Task 3: Answer and Question
  Selection for Question Answering on Arabic and English Fora}}, See
  \citeN{semeval:16}, \bibinfo{pages}{896--903}.
\newblock


\bibitem[\protect\citeauthoryear{Brown, Cocke, {Della Pietra}, {Della Pietra},
  Jelinek, Lafferty, Mercer, and Roossin}{Brown et~al\mbox{.}}{1990}]%
        {Brown:90}
\bibfield{author}{\bibinfo{person}{{Peter F.} Brown}, \bibinfo{person}{John
  Cocke}, \bibinfo{person}{{Stephen A.} {Della Pietra}},
  \bibinfo{person}{{Vicent J.} {Della Pietra}}, \bibinfo{person}{Frederick
  Jelinek}, \bibinfo{person}{{John D.} Lafferty}, \bibinfo{person}{{Robert L.}
  Mercer}, {and} \bibinfo{person}{{Paul S.} Roossin}.}
  \bibinfo{year}{1990}\natexlab{}.
\newblock \showarticletitle{{A Statistical Approach to Machine Translation}}.
\newblock \bibinfo{journal}{{\em Computational Linguistics\/}}
  \bibinfo{volume}{16}, \bibinfo{number}{2} (\bibinfo{year}{1990}),
  \bibinfo{pages}{79--85}.
\newblock


\bibitem[\protect\citeauthoryear{Cao, Duan, Lin, Yu, and Hon}{Cao
  et~al\mbox{.}}{2008}]%
        {Cao:2008:RQU:1367497.1367509}
\bibfield{author}{\bibinfo{person}{Yunbo Cao}, \bibinfo{person}{Huizhong Duan},
  \bibinfo{person}{Chin-Yew Lin}, \bibinfo{person}{Yong Yu}, {and}
  \bibinfo{person}{Hsiao-Wuen Hon}.} \bibinfo{year}{2008}\natexlab{}.
\newblock \showarticletitle{{Recommending Questions Using the Mdl-based Tree
  Cut Model}}. In \bibinfo{booktitle}{{\em Proceedings of WWW '08}}.
  \bibinfo{address}{New York, New York, USA}, \bibinfo{pages}{81--90}.
\newblock


\bibitem[\protect\citeauthoryear{Darwish and Mubarak}{Darwish and
  Mubarak}{2016}]%
        {DARWISH16.164}
\bibfield{author}{\bibinfo{person}{Kareem Darwish} {and} \bibinfo{person}{Hamdy
  Mubarak}.} \bibinfo{year}{2016}\natexlab{}.
\newblock \showarticletitle{{Farasa: A New Fast and Accurate Arabic Word
  Segmenter}}. In \bibinfo{booktitle}{{\em Proceedings of LREC}}.
  \bibinfo{address}{Portoro\v{z}, Slovenia}.
\newblock


\bibitem[\protect\citeauthoryear{{Dos Santos}, Barbosa, Bogdanova, and
  Zadrozny}{{Dos Santos} et~al\mbox{.}}{2015}]%
        {dossantos-EtAl:2015:ACL-IJCNLP}
\bibfield{author}{\bibinfo{person}{Cicero {Dos Santos}},
  \bibinfo{person}{Luciano Barbosa}, \bibinfo{person}{Dasha Bogdanova}, {and}
  \bibinfo{person}{Bianca Zadrozny}.} \bibinfo{year}{2015}\natexlab{}.
\newblock \showarticletitle{{Learning Hybrid Representations to Retrieve
  Semantically Equivalent Questions}}. In \bibinfo{booktitle}{{\em Proceedings
  of ACL '15}}. \bibinfo{address}{Beijing, China}, \bibinfo{pages}{694--699}.
\newblock


\bibitem[\protect\citeauthoryear{Filice, Croce, Moschitti, and Basili}{Filice
  et~al\mbox{.}}{2016}]%
        {filice-EtAl:2016:SemEval}
\bibfield{author}{\bibinfo{person}{Simone Filice}, \bibinfo{person}{Danilo
  Croce}, \bibinfo{person}{Alessandro Moschitti}, {and}
  \bibinfo{person}{Roberto Basili}.} \bibinfo{year}{2016}\natexlab{}.
\newblock \showarticletitle{{KeLP at SemEval-2016 Task 3: Learning Semantic
  Relations between Questions and Answers}}, See \citeN{semeval:16},
  \bibinfo{pages}{1116--1123}.
\newblock


\bibitem[\protect\citeauthoryear{Forner, Pe{\~{n}}as, Agirre, Alegria, Forascu,
  Moreau, Osenova, Prokopidis, Rocha, Sacaleanu, Sutcliffe, and Sang}{Forner
  et~al\mbox{.}}{2008}]%
        {clef/FornerPAAFMOPRSSS08}
\bibfield{author}{\bibinfo{person}{Pamela Forner}, \bibinfo{person}{Anselmo
  Pe{\~{n}}as}, \bibinfo{person}{Eneko Agirre}, \bibinfo{person}{I{\~{n}}aki
  Alegria}, \bibinfo{person}{Corina Forascu}, \bibinfo{person}{Nicolas Moreau},
  \bibinfo{person}{Petya Osenova}, \bibinfo{person}{Prokopis Prokopidis},
  \bibinfo{person}{Paulo Rocha}, \bibinfo{person}{Bogdan Sacaleanu},
  \bibinfo{person}{Richard F.~E. Sutcliffe}, {and} \bibinfo{person}{Erik F.
  Tjong~Kim Sang}.} \bibinfo{year}{2008}\natexlab{}.
\newblock \showarticletitle{{Overview of the CLEF 2008 Multilingual Question
  Answering Track}}, See \citeN{clef:08}, \bibinfo{pages}{262--295}.
\newblock


\bibitem[\protect\citeauthoryear{Giordani and Moschitti}{Giordani and
  Moschitti}{2010}]%
        {Giordani:2009:SMN:2174299.2174323}
\bibfield{author}{\bibinfo{person}{Alessandra Giordani} {and}
  \bibinfo{person}{Alessandro Moschitti}.} \bibinfo{year}{2010}\natexlab{}.
\newblock \showarticletitle{Semantic Mapping Between Natural Language Questions
  and SQL Queries via Syntactic Pairing}. In \bibinfo{booktitle}{{\em
  Proceedings of NLDB'09}}. \bibinfo{address}{Saarbr\"{u}cken, Germany},
  \bibinfo{pages}{207--221}.
\newblock
\showISBNx{3-642-12549-2, 978-3-642-12549-2}


\bibitem[\protect\citeauthoryear{Giordani and Moschitti}{Giordani and
  Moschitti}{2012}]%
        {DBLP:conf/coling/GiordaniM12}
\bibfield{author}{\bibinfo{person}{Alessandra Giordani} {and}
  \bibinfo{person}{Alessandro Moschitti}.} \bibinfo{year}{2012}\natexlab{}.
\newblock \showarticletitle{Translating Questions to {SQL} Queries with
  Generative Parsers Discriminatively Reranked}. In \bibinfo{booktitle}{{\em
  Proceedings of {COLING} '12}}. \bibinfo{address}{Mumbai, India},
  \bibinfo{pages}{401--410}.
\newblock


\bibitem[\protect\citeauthoryear{Guzm{\'a}n, M{\`a}rquez, and Nakov}{Guzm{\'a}n
  et~al\mbox{.}}{2016}]%
        {ACL2016:MTE-NN-cQA}
\bibfield{author}{\bibinfo{person}{Francisco Guzm{\'a}n},
  \bibinfo{person}{Llu{\'i}s M{\`a}rquez}, {and} \bibinfo{person}{Preslav
  Nakov}.} \bibinfo{year}{2016}\natexlab{}.
\newblock \showarticletitle{Machine Translation Evaluation Meets Community
  Question Answering}. In \bibinfo{booktitle}{{\em Proceedings of ACL~'16}}.
  \bibinfo{address}{Berlin, Germany}, \bibinfo{pages}{460--466}.
\newblock


\bibitem[\protect\citeauthoryear{Hartrumpf, Gl{{\"o}}ckner, and
  Leveling}{Hartrumpf et~al\mbox{.}}{2008}]%
        {Hartrumpf:08}
\bibfield{author}{\bibinfo{person}{Sven Hartrumpf}, \bibinfo{person}{Ingo
  Gl{{\"o}}ckner}, {and} \bibinfo{person}{Johannes Leveling}.}
  \bibinfo{year}{2008}\natexlab{}.
\newblock \showarticletitle{{University of Hagen at QA@CLEF 2008: Efficient
  Question Answering with Question Decomposition and Multiple Answer Streams}},
  See \citeN{clef:08}, \bibinfo{pages}{421--428}.
\newblock


\bibitem[\protect\citeauthoryear{Jeon, Croft, and Lee}{Jeon et~al\mbox{.}}{}]%
        {Jeon:2005:FSQ:1099554.1099572}
\bibfield{author}{\bibinfo{person}{Jiwoon Jeon}, \bibinfo{person}{W.~Bruce
  Croft}, {and} \bibinfo{person}{Joon~Ho Lee}.}
\newblock \showarticletitle{{Finding Similar Questions in Large Question and
  Answer Archives}}. In \bibinfo{booktitle}{{\em Proceedings of CIKM '05}}.
  \bibinfo{address}{Bremen, Germany}, \bibinfo{pages}{84--90}.
\newblock


\bibitem[\protect\citeauthoryear{Ji, Xu, Wang, and He}{Ji
  et~al\mbox{.}}{2012}]%
        {ji2012question}
\bibfield{author}{\bibinfo{person}{Zongcheng Ji}, \bibinfo{person}{Fei Xu},
  \bibinfo{person}{Bin Wang}, {and} \bibinfo{person}{Ben He}.}
  \bibinfo{year}{2012}\natexlab{}.
\newblock \showarticletitle{{Question-Answer Topic Model for Question Retrieval
  in Community Question Answering}}. In \bibinfo{booktitle}{{\em Proceedings of
  CIKM '12}}. \bibinfo{address}{Maui, Hawaii, USA},
  \bibinfo{pages}{2471--2474}.
\newblock


\bibitem[\protect\citeauthoryear{Lin and Kuo}{Lin and Kuo}{2010}]%
        {Lin:10}
\bibfield{author}{\bibinfo{person}{Chuan-Jie Lin} {and} \bibinfo{person}{Yu-Min
  Kuo}.} \bibinfo{year}{2010}\natexlab{}.
\newblock \showarticletitle{{Description of the NTOU Complex QA System}}, See
  \citeN{ntcir:8}, \bibinfo{pages}{47--54}.
\newblock


\bibitem[\protect\citeauthoryear{Luong, Pham, and Manning}{Luong
  et~al\mbox{.}}{2015}]%
        {luong-pham-manning:2015:VSM-NLP}
\bibfield{author}{\bibinfo{person}{Thang Luong}, \bibinfo{person}{Hieu Pham},
  {and} \bibinfo{person}{Christopher~D. Manning}.}
  \bibinfo{year}{2015}\natexlab{}.
\newblock \showarticletitle{Bilingual Word Representations with Monolingual
  Quality in Mind}. In \bibinfo{booktitle}{{\em Proceedings of the 1st Workshop
  on Vector Space Modeling for NLP}}. \bibinfo{address}{Denver, Colorado, USA},
  \bibinfo{pages}{151--159}.
\newblock


\bibitem[\protect\citeauthoryear{Mikolov, Yih, and Zweig}{Mikolov
  et~al\mbox{.}}{2013}]%
        {mikolov-yih-zweig:2013:NAACL-HLT}
\bibfield{author}{\bibinfo{person}{Tomas Mikolov}, \bibinfo{person}{Wen-tau
  Yih}, {and} \bibinfo{person}{Geoffrey Zweig}.}
  \bibinfo{year}{2013}\natexlab{}.
\newblock \showarticletitle{Linguistic Regularities in Continuous Space Word
  Representations}. In \bibinfo{booktitle}{{\em Proceedings of NAACL-HLT~'13}}.
  \bibinfo{address}{Atlanta, Georgia, USA}, \bibinfo{pages}{746--751}.
\newblock


\bibitem[\protect\citeauthoryear{Mitamura, Shima, Sakai, Kando, Mori, Takeda,
  Lin, Lin, and Lee}{Mitamura et~al\mbox{.}}{2010}]%
        {mitamura:10}
\bibfield{author}{\bibinfo{person}{Teruko Mitamura}, \bibinfo{person}{Hideki
  Shima}, \bibinfo{person}{Tetsuya Sakai}, \bibinfo{person}{Noriko Kando},
  \bibinfo{person}{Tatsunori Mori}, \bibinfo{person}{Koichi Takeda},
  \bibinfo{person}{Ruihua Lin, Chin-Yew~Song}, \bibinfo{person}{Chuan-Jie Lin},
  {and} \bibinfo{person}{Cheng-Wei Lee}.} \bibinfo{year}{2010}\natexlab{}.
\newblock \showarticletitle{{Overview of the NTCIR-8 ACLIA Tasks: Advanced
  Cross-Lingual Information Access}}, See \citeN{ntcir:8},
  \bibinfo{pages}{15--24}.
\newblock


\bibitem[\protect\citeauthoryear{Nakov, M\`{a}rquez, and Guzm\'{a}n}{Nakov
  et~al\mbox{.}}{2016a}]%
        {nakov-marquez-guzman:2016:EMNLP2016}
\bibfield{author}{\bibinfo{person}{Preslav Nakov}, \bibinfo{person}{Llu\'{i}s
  M\`{a}rquez}, {and} \bibinfo{person}{Francisco Guzm\'{a}n}.}
  \bibinfo{year}{2016}\natexlab{a}.
\newblock \showarticletitle{It Takes Three to Tango: Triangulation Approach to
  Answer Ranking in Community Question Answering}. In \bibinfo{booktitle}{{\em
  Proceedings of EMNLP '16}}. \bibinfo{address}{Austin, Texas, USA},
  \bibinfo{pages}{1586--1597}.
\newblock


\bibitem[\protect\citeauthoryear{Nakov, M\`{a}rquez, Moschitti, Magdy, Mubarak,
  Freihat, Glass, and Randeree}{Nakov et~al\mbox{.}}{2016b}]%
        {nakov-EtAl:2016:SemEval2}
\bibfield{author}{\bibinfo{person}{Preslav Nakov}, \bibinfo{person}{Llu\'{i}s
  M\`{a}rquez}, \bibinfo{person}{Alessandro Moschitti}, \bibinfo{person}{Walid
  Magdy}, \bibinfo{person}{Hamdy Mubarak}, \bibinfo{person}{{abed Alhakim}
  Freihat}, \bibinfo{person}{Jim Glass}, {and} \bibinfo{person}{Bilal
  Randeree}.} \bibinfo{year}{2016}\natexlab{b}.
\newblock \showarticletitle{{SemEval-2016 Task 3: Community Question
  Answering}}, See \citeN{semeval:16}, \bibinfo{pages}{525--545}.
\newblock


\bibitem[\protect\citeauthoryear{Nicosia, Filice, Barr\'{o}n-Cede\~{n}o, Saleh,
  Mubarak, Gao, Nakov, Da~San~Martino, Moschitti, Darwish, M\`{a}rquez, Joty,
  and Magdy}{Nicosia et~al\mbox{.}}{}]%
        {nicosia-EtAl:2015:SemEval}
\bibfield{author}{\bibinfo{person}{Massimo Nicosia}, \bibinfo{person}{Simone
  Filice}, \bibinfo{person}{Alberto Barr\'{o}n-Cede\~{n}o},
  \bibinfo{person}{Iman Saleh}, \bibinfo{person}{Hamdy Mubarak},
  \bibinfo{person}{Wei Gao}, \bibinfo{person}{Preslav Nakov},
  \bibinfo{person}{Giovanni Da~San~Martino}, \bibinfo{person}{Alessandro
  Moschitti}, \bibinfo{person}{Kareem Darwish}, \bibinfo{person}{Llu\'{i}s
  M\`{a}rquez}, \bibinfo{person}{Shafiq Joty}, {and} \bibinfo{person}{Walid
  Magdy}.}
\newblock \showarticletitle{{QCRI: Answer Selection for Community Question
  Answering - Experiments for {A}rabic and {E}nglish}}. In
  \bibinfo{booktitle}{{\em Proceedings of SemEval~'15}}.
  \bibinfo{address}{Denver, Colorado, USA}, \bibinfo{pages}{203--209}.
\newblock


\bibitem[\protect\citeauthoryear{Ren, Ji, and Wan}{Ren et~al\mbox{.}}{2010}]%
        {Ren:10}
\bibfield{author}{\bibinfo{person}{Han Ren}, \bibinfo{person}{Donghong Ji},
  {and} \bibinfo{person}{Jing Wan}.} \bibinfo{year}{2010}\natexlab{}.
\newblock \showarticletitle{{WHU Question Answering System at NTCIR-8 ACLIA
  Task}}, See \citeN{ntcir:8}, \bibinfo{pages}{31--36}.
\newblock


\bibitem[\protect\citeauthoryear{Severyn and Moschitti}{Severyn and
  Moschitti}{2012}]%
        {sigir12}
\bibfield{author}{\bibinfo{person}{Aliaksei Severyn} {and}
  \bibinfo{person}{Alessandro Moschitti}.} \bibinfo{year}{2012}\natexlab{}.
\newblock \showarticletitle{{Structural Relationships for Large-scale Learning
  of Answer Re-Ranking}}. In \bibinfo{booktitle}{{\em Proceedings of SIGIR
  '12}}. \bibinfo{address}{Portland, Oregon, USA}, \bibinfo{pages}{741--750}.
\newblock


\bibitem[\protect\citeauthoryear{Soboroff, Griffitt, and Strassel}{Soboroff
  et~al\mbox{.}}{2016}]%
        {Soboroff:2016}
\bibfield{author}{\bibinfo{person}{Ian Soboroff}, \bibinfo{person}{Kira
  Griffitt}, {and} \bibinfo{person}{Stephanie Strassel}.}
  \bibinfo{year}{2016}\natexlab{}.
\newblock \showarticletitle{The BOLT IR Test Collections of Multilingual
  Passage Retrieval from Discussion Forums}. In \bibinfo{booktitle}{{\em
  Proceedings of SIGIR '16}}. \bibinfo{address}{New York, New York, USA},
  \bibinfo{pages}{713--716}.
\newblock
\showISBNx{978-1-4503-4069-4}


\bibitem[\protect\citeauthoryear{Socher, Bauer, Manning, and Andrew~Y.}{Socher
  et~al\mbox{.}}{2013}]%
        {socher-EtAl:2013:ACL2013}
\bibfield{author}{\bibinfo{person}{Richard Socher}, \bibinfo{person}{John
  Bauer}, \bibinfo{person}{Christopher~D. Manning}, {and} \bibinfo{person}{Ng
  Andrew~Y.}} \bibinfo{year}{2013}\natexlab{}.
\newblock \showarticletitle{Parsing with Compositional Vector Grammars}. In
  \bibinfo{booktitle}{{\em Proceedings of ACL '13}}. \bibinfo{address}{Sofia,
  Bulgaria}, \bibinfo{pages}{455--465}.
\newblock


\bibitem[\protect\citeauthoryear{Tiedemann}{Tiedemann}{2012}]%
        {Tiedemann:12}
\bibfield{author}{\bibinfo{person}{J\"{o}rg Tiedemann}.}
  \bibinfo{year}{2012}\natexlab{}.
\newblock \showarticletitle{Parallel Data, Tools and Interfaces in {OPUS}}. In
  \bibinfo{booktitle}{{\em Proceedings of LREC'12}}.
  \bibinfo{address}{Istanbul, Turkey}.
\newblock


\bibitem[\protect\citeauthoryear{Ture and Boschee}{Ture and Boschee}{2016}]%
        {ture-boschee:2016:EMNLP2016}
\bibfield{author}{\bibinfo{person}{Ferhan Ture} {and}
  \bibinfo{person}{Elizabeth Boschee}.} \bibinfo{year}{2016}\natexlab{}.
\newblock \showarticletitle{Learning to Translate for Multilingual Question
  Answering}. In \bibinfo{booktitle}{{\em Proceedings of EMNLP '16}}.
  \bibinfo{address}{Austin, Texas, USA}, \bibinfo{pages}{573--584}.
\newblock


\bibitem[\protect\citeauthoryear{Upadhyay, Faruqui, Dyer, and Roth}{Upadhyay
  et~al\mbox{.}}{2016}]%
        {upadhyay-EtAl:2016:P16-1}
\bibfield{author}{\bibinfo{person}{Shyam Upadhyay}, \bibinfo{person}{Manaal
  Faruqui}, \bibinfo{person}{Chris Dyer}, {and} \bibinfo{person}{Dan Roth}.}
  \bibinfo{year}{2016}\natexlab{}.
\newblock \showarticletitle{Cross-lingual Models of Word Embeddings: An
  Empirical Comparison}. In \bibinfo{booktitle}{{\em Proceedings of ACL}}.
  \bibinfo{address}{Berlin, Germany}, \bibinfo{pages}{1661--1670}.
\newblock


\bibitem[\protect\citeauthoryear{Wang, Ming, and Chua}{Wang
  et~al\mbox{.}}{2009}]%
        {wang2009syntactic}
\bibfield{author}{\bibinfo{person}{Kai Wang}, \bibinfo{person}{Zhaoyan Ming},
  {and} \bibinfo{person}{Tat-Seng Chua}.} \bibinfo{year}{2009}\natexlab{}.
\newblock \showarticletitle{{A Syntactic Tree Matching Approach to Finding
  Similar Questions in Community-based QA Services}}. In
  \bibinfo{booktitle}{{\em Proceedings of SIGIR '09}}.
  \bibinfo{address}{Boston, Massachusetts, USA}, \bibinfo{pages}{187--194}.
\newblock


\bibitem[\protect\citeauthoryear{Zhang, Wu, Wu, Li, and Zhou}{Zhang
  et~al\mbox{.}}{2014}]%
        {zhang2014question}
\bibfield{author}{\bibinfo{person}{Kai Zhang}, \bibinfo{person}{Wei Wu},
  \bibinfo{person}{Haocheng Wu}, \bibinfo{person}{Zhoujun Li}, {and}
  \bibinfo{person}{Ming Zhou}.} \bibinfo{year}{2014}\natexlab{}.
\newblock \showarticletitle{{Question Retrieval with High Quality Answers in
  Community Question Answering}}. In \bibinfo{booktitle}{{\em Proceedings of
  CIKM '14}}. \bibinfo{address}{Shangai, China}, \bibinfo{pages}{371--380}.
\newblock


\bibitem[\protect\citeauthoryear{Zhou, Cai, Zhao, and Liu}{Zhou
  et~al\mbox{.}}{2011}]%
        {zhou2011phrase}
\bibfield{author}{\bibinfo{person}{Guangyou Zhou}, \bibinfo{person}{Li Cai},
  \bibinfo{person}{Jun Zhao}, {and} \bibinfo{person}{Kang Liu}.}
  \bibinfo{year}{2011}\natexlab{}.
\newblock \showarticletitle{{Phrase-based translation model for question
  retrieval in community question answer archives}}. In
  \bibinfo{booktitle}{{\em Proceedings of ACL}}. \bibinfo{address}{Portland,
  Oregon, USA}, \bibinfo{pages}{653--662}.
\newblock


\end{thebibliography}

\end{document}